\documentclass{bmvc2k}
\usepackage{amssymb}
\usepackage{amsmath}
\usepackage{multirow}
\usepackage{blindtext}
\usepackage{floatrow}

\title{Attribute Adaptive Margin Softmax Loss using Privileged Information}

\addauthor{Seyed Mehdi Iranmanesh}{seiranmanesh@mix.wvu.edu}{1}
\addauthor{Ali Dabouei}{ad0046@mix.wvu.edu}{1}
\addauthor{Nasser M. Nasrabadi}{nasser.nasrabadi@mail.wvu.edu}{1}

\addinstitution{
 West Virginia University\\
 West Virginia, USA\\
}

\runninghead{Iranmanesh, Dabouei, Nasrabadi}{Attribute Adaptive Margin Softmax}

\begin{document}

\maketitle

\begin{abstract}
We present a novel framework to exploit privileged information for recognition which is provided only during the training phase. Here, we focus on recognition task where images are provided as the main view and soft biometric traits (attributes) are provided as the privileged data (only available during training phase). We demonstrate that more discriminative feature space can be learned by enforcing a deep network to adjust adaptive margins between classes utilizing attributes. This tight constraint also effectively reduces the class imbalance inherent in the local data neighborhood, thus carving more balanced class boundaries locally and using feature space more efficiently. Extensive experiments are performed on five different datasets and the results show the superiority of our method compared to the state-of-the-art models in both tasks of face recognition and person re-identification.
\end{abstract}


\section{Introduction}

A multimodal recognition system with multiple modalities, such as the face, fingerprint, and iris, is expected to be more reliable and accurate due to the utilization of different sources of information. However, acquisition of this information is costly and a tedious task which can affect the popularity and ease of using multimodal recognition systems. On the other hand, there are some informative traits, such as age, gender, ethnicity, race, and hair color, which are not distinctive enough for the sake of recognition, but still can act as complementary information to other primary information, such as face and fingerprint. These traits, which are known as soft biometrics, can improve recognition algorithms performance.

The design of a recognition system comprises two major phases, namely training and testing. However, in some cases, there are extra information which is only available during the training phase and is missing during the testing phase. In other words, the training data is augmented with some extra auxiliary information. For example, in object recognition, the labeled images may be annotated with texts which can provide semantic information about the object, or any other extra knowledge, such as the boundary information of an object which determines the exact location of a specific object~\cite{3}. This extra information can be regarded as an auxiliary to the primary modality of the data. Unlike the domain adaptation and transfer learning problems in which the data is similar in both the source and target domains but statistically different~\cite{4, 5}, here, the available data in the source domain has an extra modality which is not available in the target domain.

		


The concept of learning using privileged information (LUPI) was introduced in~\cite{9}, for the first time, under the context of SVM+. In their methodology, the auxiliary data, or privileged information, was employed to predict the slack variables of an SVM classifier~\cite{62}. Since then, the idea of using privileged information has been vastly investigated in the literature and has been applied to different applications and contexts, e.g., object localization~\cite{21}, facial feature detection~\cite{10}, and metric learning~\cite{22}. A framework was introduced in~\cite{3} to embed the main data into the latent feature subspace such that the mutual information between the embedded data and auxiliary data is maximized.



Softmax loss is widely used in training CNN features~\cite{34}, which is specified as a combination of the last fully connected layer, a softmax function and a crossentropy loss~\cite{liu2016large}. However, features through softmax loss are learned with limited discriminative power. To address the limitation, various supervision objectives have been proposed to enhance the discriminativeness of the learned features, such as contrastive loss~\cite{sun2014deep}, triplet loss~\cite{schroff2015facenet}, center loss~\cite{wen2016discriminative}. In contrast to most of the other loss functions which use Euclidean margin,~\cite{liu2017sphereface,wang2018cosface,deng2019arcface} showed the effectiveness of angular margin to squeeze each class. However, these methods have an implicit hypothesis that all the classes have sufficient samples
to describe their distributions, so that a constant margin is enough to equally squeeze each intra-class variations. However this is not the case in many public unbalanced datasets.

\begin{figure}
\begin{tabular}{cc}
\bmvaHangBox{\fbox{\includegraphics[width=6.7cm]{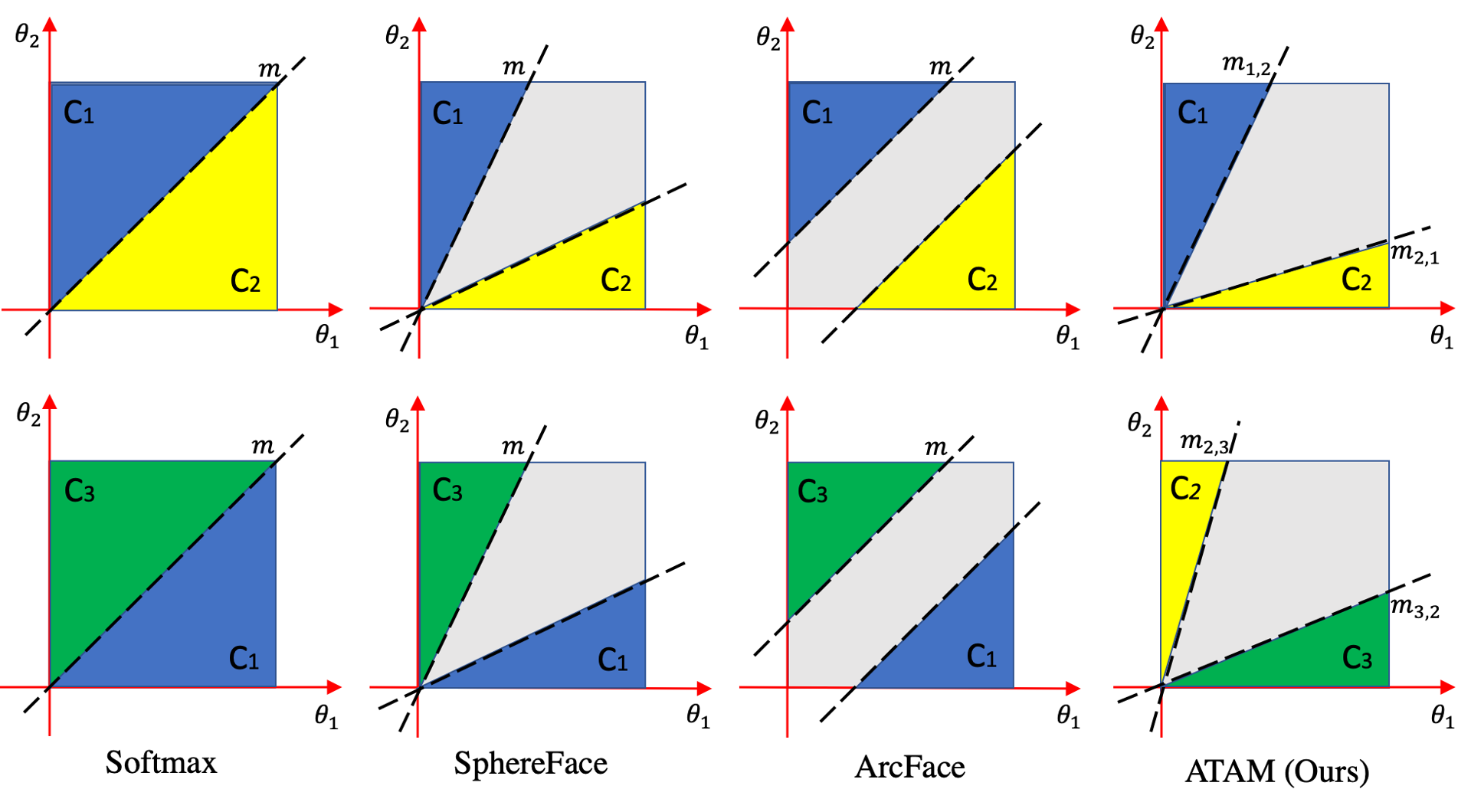}}}&
\bmvaHangBox{\fbox{\includegraphics[width=4.5cm]{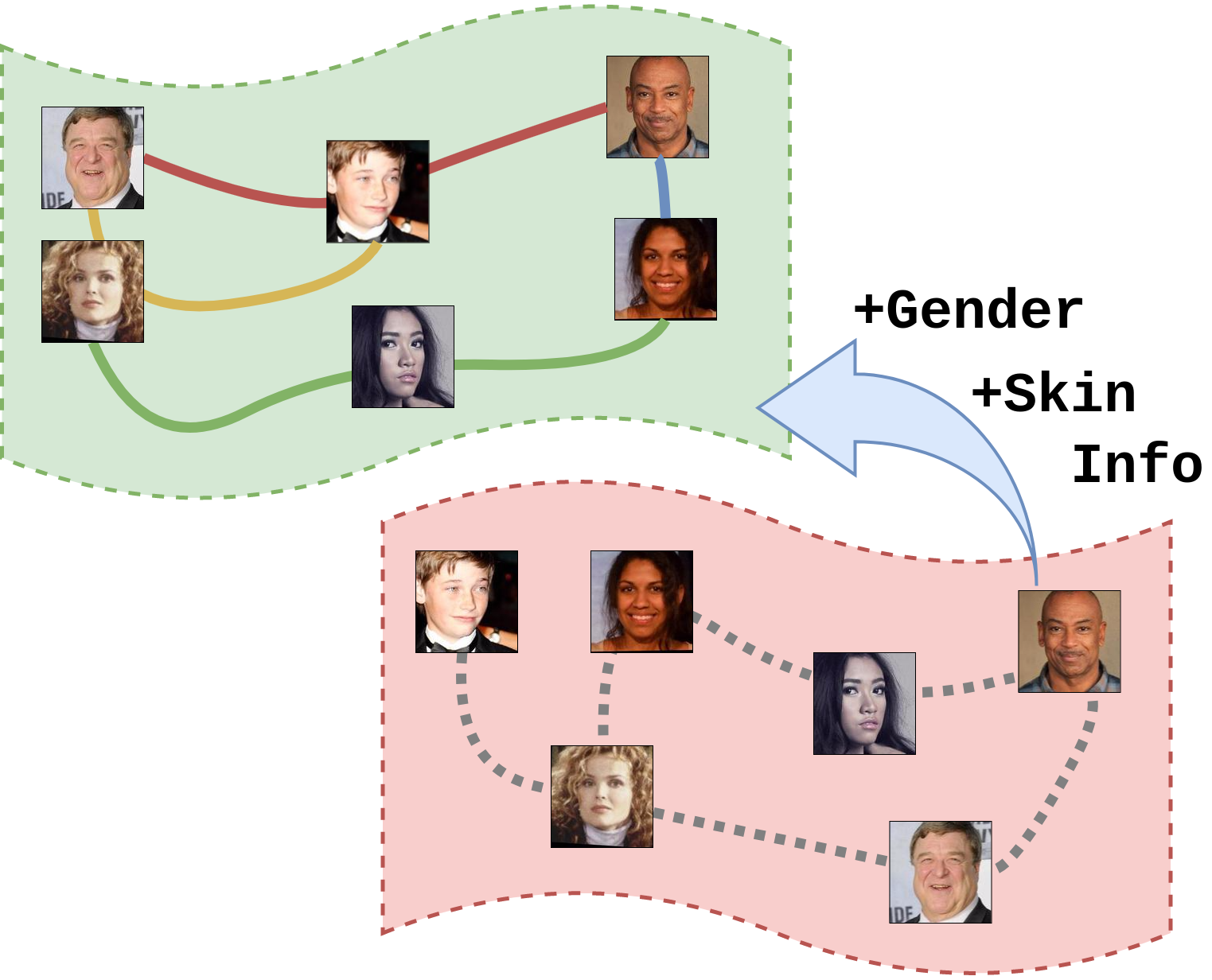}}}\\
(a)&(b)
\end{tabular}
\caption{(a) Decision margins of different loss functions for three different classes C1, C2, and C3 (in blue, yellow, and green, respectively). The dashed line represents the decision boundary, and the grey areas are the decision margins; (b) The proposed framework employs attributes information to improve the semantic correlation of the faces in the embedding space. Top and bottom figures illustrate the deep embedding of faces using the proposed and conventional methods, respectively. }
\label{fig:teaser}
\end{figure}

Our proposed training scheme has the ability to transfer information from the auxiliary data and help the network to learn a more discriminative features regarding the primary data (see Fig.~\ref{fig:teaser} (b)). To this end, we propose a novel loss
function, Attribute Adaptive Margin Softmax Loss (ATAM), to adaptively find the appropriate margins utilizing attributes during training phase. Specifically, we make the margin $m$ particular and learnable and directly train the network to find the adaptive margins. We show its important applications to face recognition and person re-identification. Note such tasks can be assessed under either closed- or open-set protocol. The open-set protocol is harder since the testing classes may be unseen from the training classes. It usually requires discriminative feature representations with built-in large margins, which are embodied in our approach.

\begin{figure}
	\begin{center}
		\includegraphics[width=1\linewidth]{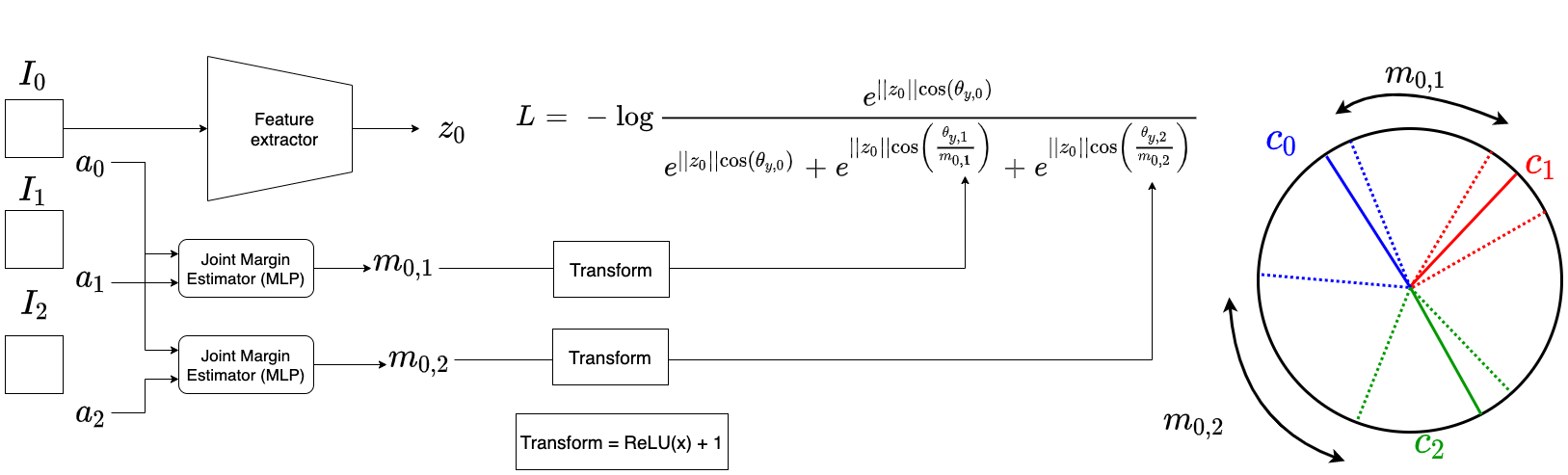}
		
	\end{center}
	\caption{Proposed ATAM loss utilizing privileged attributes. }
	\label{fig:network}
\end{figure}

Our method is motivated by the observation that the minority classes often encompasses very few samples with high degree of visual variability. The scarcity and high variability makes the neighborhood of these samples easy to be invaded by other imposter nearest neighbors. To this end, we propose to learn an embedding utilizing attributes along with a novel loss function to ameliorate the invasion. For the purpose of assessment, we study the problem of face recognition and person re-identification using paired image-attribute data, while the attributes (i.e., soft biometrics) are only available during the training phase (see Fig.~\ref{fig:network}).

The major contributions of the paper are: i) rewriting the angular softmax loss using a set of inter-class margins, ii) proposing a framework enhanced by auxiliary information to learn these margins simultaneously during the training, iii) providing an example for the auxiliary information by means of the discrepancy between the attributes. The performance of the proposed scheme is evaluated on five different datasets, namely MegaFace~\cite{kemelmacher2016megaface}, YTF~\cite{wolf2011face}, LFW~\cite{huang2008labeled}, Market-1501 ~\cite{zheng2015scalable}, and DukeMTMC-reID~\cite{zheng2017unlabeled} datasets. On both
of the mentioned tasks, we demonstrate the superiority of ATAM loss with performance on par with the state of the art.

\section{Methodology}
\label{sec:method}
In this section we detail our methodology. First we revisit the A-Softmax~\cite{liu2017sphereface} which maps faces on the hyperspace manifold. Then we present our proposed attribute adaptive margin Softmax loss which exploits the attributes to learn more discriminative feature space. 

\subsection{A-Softmax}
Lets begin with the most widely used Softmax loss. The Softmax loss maximizes the posterior probability of each class to separate features of different classes. Its formulation is provided as follows:

\begin{equation}
L_s = -\dfrac{1}{N}\sum_{i=1}^{N} \log \dfrac{e ^ {W^T_{y_i}z_i+b_{y_i}}}{\sum_{j=1}^{M} e ^ {W^T_{j} z_i+b_j}} \;,
\label{Eq-1}
\end{equation}

\noindent where $W_j \in \mathbb{R}^d$ is the weights of last layer of class $j$ with $d$ dimension and $b_j \in \mathbb{R}$ is the bias term. $z_i \in \mathbb{R}^d$ is the learned feature of sample $i$, and $y_i$ is the ground truth class label. $N$ and $M$ are the number of samples and classes, respectively. The inherent angular distribution of learned deep features by Softmax loss suggests using cosine distance as the metric instead of using Euclidean distance~\cite{liu2017sphereface}. Modified Softmax loss normalizes the weights $||W_i|| =1$ and zero the biases as follows:

\begin{equation}
L_m = -\dfrac{1}{N}\sum_{i=1}^{N} \log \dfrac{e ^ {||z_i||\cos(\theta_{y_i,i})}}{\sum_{j=1}^{M} e^ {||z_i||\cos(\theta_{j,i})}}.
\label{Eq-2}
\end{equation}

Given a query point, the model compares its angle with weights of different classes and select the one with minimum angle. Although the features learned using modified Softmax have angular boundary, they are not necessarily discriminative enough. SphereFace~\cite{liu2017sphereface} proposed a natural way to produce the angular margins through an A-Softmax loss. The angle between a query point and target class is multiplied by the margin parameter $m$:

\begin{equation}
L_{a-s} = -\dfrac{1}{N}\sum_{i=1}^{N} \log \dfrac{e ^ {||z_i||\psi(\theta_{y_i,i})}}{ e^ {||z_i||\psi(\theta_{y_i,i})}+ \sum_{j\neq{y_i}} e^ {||z_i||\cos(\theta_{j,i})}} \;,
\label{Eq-3}
\end{equation}

\noindent where $\psi(\theta_{y_i},i)$ is a monotonically decreasing angle function and defined as $(-1)^ k \cos(m\theta_{y_i}, i) - 2k$, and $\theta_{y_i}, i \in [\dfrac{k\pi}{m},\dfrac{(k+1)\pi}{m}]$, $k \in [0,m-1] $ to compensate for the limitation of $\theta_{y_i}, i$ in $\cos(m\theta_{y_i}, i)$. While A-Softmax loss manually tune $m$ to squeeze the intra-class variation and increase the angular distance between different classes, it fails to consider the feature distribution in the hyperspace manifold. Other works such as CosFace~\cite{wang2018cosface} and ArcFace~\cite{deng2019arcface} also have similar assumption and consider same distributions for different classes. Thus, they fail to exploit the holistic feature space.~\cite{duan2019uniformface,liu2019adaptiveface} tried to address this limitation by designing a new loss function and constructing a weighted combination with A-Softmax loss function. 

Here, we propose a natural way to learn and tune $m$ in an end-to-end fashion in order to make the the holistic feature space more discriminative and consider the inter-class space between different classes via their attributes.   

\subsection{Attribute Adaptive Margin Softmax Loss }\label{subsection:atam}

Given an Image $I$ and its specific attribute set $a$, we propose to learn adaptive margins which discriminate the holistic feature space and reflects the characteristics of corresponding attribute in the image. If there are $k$ attributes $\tfrac{k \times (k-1)}{2}$ margins can be learned in the feature space. The proposed model architecture is composed of a feature extraction branch (CNN) combined with a small network from attributes. The attribute network is a MLP network which is responsible for learning margins. Concretely, attribute sets $a_j, a_{y_i}$ (each of which has dimensionality of $k$) related to the specific samples from classes $j,y_i$, respectively, are fused with simple concatenation and further fed into three subsequent $FC$ layers (MLP network) to obtain the specific margin ($m_{j,y_i}$ or $m_{y_i,j}$). 

The margin $m$ is usually set manually in SphereFace~\cite{liu2017sphereface}, and also kept constant during the training phase. In order to address the problem mentioned above and have a better holistic feature space, we propose to have a learnable attribute-specific parameter $m$. The Eq.~\ref{Eq-3} can be modified as follows:

\begin{equation}
L_{ATAM} = -\dfrac{1}{N}\sum_{i=1}^{N} \log \dfrac{e ^ {||z_i||\cos(\theta_{y_i,i})}}{ e^ {||z_i||\cos(\theta_{y_i,i})} +\sum_{j\neq{y_i}} e^ {||z_i||\cos(\dfrac{\theta_{j,i}}{m_{j,y_i}})}} \;,
\label{Eq-6}
\end{equation}

\noindent where $m_{j,y_i}$ is the score that is provided by the attribute network (MLP network) with the input $[a_j,a_{y_i}]$ (where $[,]$ is the concatenation operation). Note that since we want the margins to be greater or equal to 1 ($m_{y_i,j} \geq 1$), we normalize the output of MLP network using $ReLU$ function. Afterwards, this positive score is added to 1 (see Fig~\ref{sec:method}). This transformation ($ReLU(m_{j,y_i})+1$) ensures that $m_{j,y_i}= 1$ in the worst case situation. Both of the CNN and MLP branches are trained jointly using Eq.~\ref{Eq-6} in an end-to-end fashion.

\subsection{Discussion}

In this subsection we compare our proposed ATAM loss with Softmax, A-Softmax~\cite{liu2017sphereface}, and ArcFace~\cite{deng2019arcface} as illustrated in Fig.~\ref{fig:teaser} (a). For simplicity of analysis, we consider three classes of C1, C2, and C3. 

Softmax decision boundary depends on both magnitudes of weight
vectors and cosine of angles, which results in an overlapping decision area (margin < 0) in the cosine space. The popular A-Softmax reduces intra-class variations and allocates equal space for each class, without considering its sample distribution. In contrast to A-Softmax which has a nonlinear angular margin, ArcFace has a constant linear angular margin throughout the whole interval. However, it still fails to capture the true distribution of each class in the holistic feature space leading to not accurate margins between classes. 


We introduce learnable attribute-specific margins {($m_{1,2}$, $m_{1,3}$, and $m_{2,3}$ in Fig.~\ref{fig:teaser} (a))} by utilizing the privileged attribute information during the training phase. This loss intrinsically takes attributes into account in a unified loss function leading to the more discriminative holistic feature space. The attribute-specific margins discriminate embedding domain into different clusters and each cluster may have one or more classes. This also embraces the multimodality of class distribution: it preserves not only locality across the same-class with different attributes but also discrimination between classes. Hence, it is capable of preserving discrimination in any local neighborhood, and forming local class boundaries with the most discriminative samples with regards to the attributes (More distant classes contain more different attributes: class C2 (yellow) is placed between class C1 (blue) and class C3 (green) due to the more discrepancy between the attributes of C1 and C3 compared to C1 and C2 in Fig.~\ref{fig:teaser} (a)).

UniformFace~\cite{duan2019uniformface} and AdaptiveFace~\cite{liu2019adaptiveface} also try to make the embedding space more efficient and discriminative. UniformFace address this by adding another loss function to A-Softmax loss with another hyperparameter in a similar fashion to old joint loss functions (i.e. different combination of contrastive loss or center loss with softmax). It disperses classes in the embedding domain uniformly which does not reflect the true underlying distributions of classes.  Adaptive loss also tackles this issue by adjusting margin $m$ adaptively. However, it also introduce extra hyperparameter $\lambda$ to jointly train additional loss with the modified A-Softmax. The additional loss function is taking the average of all classes' margins leading to overlooking specific margin for each class.

\begin{table}[]
\centering
\small
\setlength\tabcolsep{2pt}
\begin{tabular}{c|ccc}
\hline
Method                    & Protocol & MF1 Rank1 & MF1 Veri. \\ \hline
Beijing FaceAll Norm 1600 & Large    & 64.80     & 67.11     \\
Google - FaceNet v8~\cite{schroff2015facenet}       & Large    & 70.49     & 86.47     \\
NTechLAB - facenx large   & Large    & 73.30     & 85.08     \\
SIATMMLAB TencentVision   & Large    & 74.20     & 87.27     \\
DeepSense V2              & Large    & 81.29     & 95.99     \\
YouTu Lab                 & Large    & 83.29     & 91.34     \\
Vocord - deepVo V3        & Large    & 91.76     & 94.96     \\
CosFace~\cite{wang2018cosface}           & Large    & 82.72     & 96.65  
\\
UniformFace~\cite{duan2019uniformface}                 & Large    & 79.98     & 95.36     \\
AdaptiveFace~\cite{liu2019adaptiveface}          & Large    & 95.02     & 95.61     \\
\hline
Softmax                   & Large    & 71.37     & 73.05     \\
SphereFace~\cite{liu2017sphereface}        & Large    & 92.24     & 93.42     \\
CosFace~\cite{wang2018cosface}           & Large    & 93.94     & 94.11     \\
ArcFace~\cite{deng2019arcface}            & Large    & 94.64     & 94.85     \\
ATAM                      & Large    & 96.51     & 97.14    
\end{tabular}
\caption{ Face identification and verification evaluation on MF1.
“Rank 1” refers to rank-1 face identification accuracy and “Veri.” refers to face verification TAR under $10 ^ {-6}$ FAR.}
\label{Table:face}
\end{table}

The proposed loss utilizes the privileged information to learn the margins that are usually set manually in other loss functions such as cosFace loss. Our method enhances the training of the feature extractor network in two ways. First, instead of defining a global margin that is constant for all the classes, it defines k(k-1)/2 margins for the pairs of classes with different attribute information. Second, we incorporate the attribute information to adaptively control the inter-class margins and regularize the distribution of the features in the embeddings. Hence, in contrast to the previous approaches that impose a global prior for the separability of the class distributions, ATAM enables the training framework to carefully exploit the auxiliary information to learn the distribution of the features based on a set of adaptive priors constructed using the local properties of inter-class relationships, i.e., relative discrepancy of attributes. After the training, we solely use the feature extractor network for the face recognition/ re-id.

\section{Experiments}
In this section we conduct extensive experiments on five datasets to demonstrate the effectiveness of our proposed method. First, we describe the implementation setup of the proposed model in subsection~\ref{imp_detail}. Eventually, we compare the proposed ATAM to other baselines for two tasks of face recognition and person re-identification in subsections~\ref{facerec-section} and~\ref{person-reidsection}, respectively.

\subsection{Implementation Details}\label{imp_detail}

We adopt ResNet~\cite{31} as the base network of our proposed model. We performed standard preprocessing on faces. MTCNN~\cite{zhang2016joint} is used to detect and align each face via five landmarks (two eyes, two mouth corners and nose) from train and test sets. Afterwards, we cropped the image into $112 \times 112$. We also normalized each pixel in
RGB images by subtracting 127.5 and then dividing by 128.

All CNN models in the experiments use the same architecture in this paper, which is a 50-layer residual network~\cite{31}. It contains four residual blocks and finally gets a 512 dimensional feature by average pooling. The networks are trained using Stochastic Gradient Descent (SGD) on TITANX GPUs and the batch size is set to fill all the GPU memory.  The initial value for the learning rate is set to 0.1 and multiplied by 0.9 in intervals of five epochs until its value is less than or equal to $10^{-6}$. All models are trained for $600K$ iterations.

\textbf{Training:} We trained our model on the refined MSCeleb-1M~\cite{guo2016ms} dataset. MS-Celeb-1M originally contained about 10M images from 100K identities. We removed the images which were far away from the class centers to enhance the quality of the training data and cleared the identities with less than 3 images to diminish
the long-tail distribution~\cite{deng2019arcface,deng2017marginal}. The refined MS-Celeb-1M dataset contained 85K identities with 3.84M images. The face images are horizontally flipped for data augmentation.

\begin{figure}
	\begin{center}
		\includegraphics[width=0.7\linewidth]{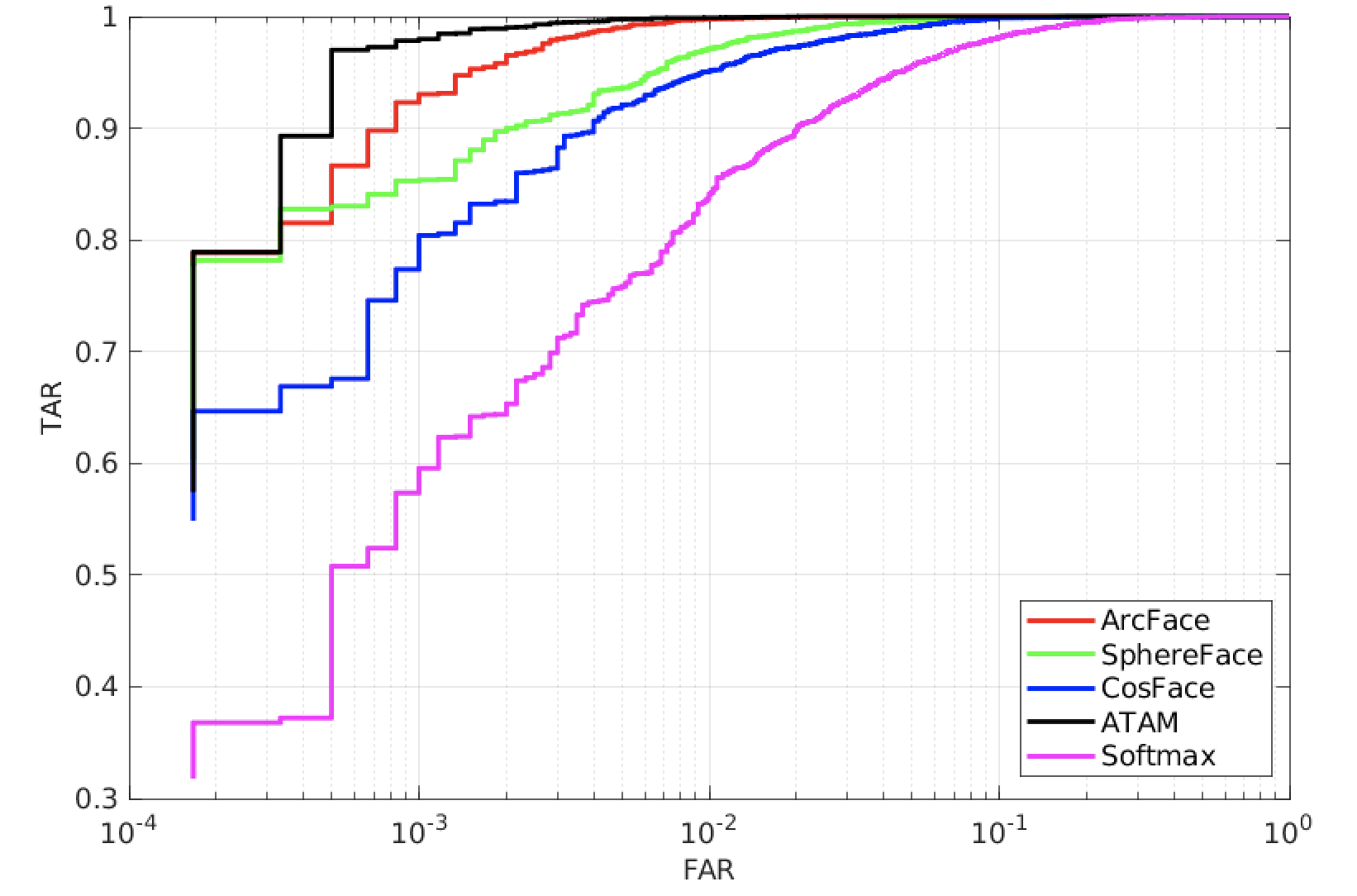}
		
	\end{center}
	\caption{ROC curves for matching face images for different methods on LFW~\cite{huang2008labeled}.}
	\label{fig:roc}
\end{figure}

\textbf{Evaluation Setup:} For each image, we extract features only from the original image as the final representation. We didn’t extract features from both the original image and the flipped one and concatenate them as the final representation. Therefore, the dimension of the final representation is 512 for each image. The score is measured by the cosine distance of two features. Eventually, verification and identification are conducted by thresholding and ranking the scores.

\subsection{Face Recognition: Overall Benchmark Comparisons}\label{facerec-section}
\textbf{Experiments on MegaFace.}

MegaFace~\cite{kemelmacher2016megaface} is one of the most challenging testing benchmark for large-scale face identification and verification, which intends to assess the performance of face recognition models at the million scale of distractors. The gallery set of MegaFace is a subset of Flickr photos, contains more than one million face images. The probe sets are two existing databases: FaceScrub~\cite{ng2014data} and FGNet. The FaceScrub dataset contains 106,863 face images of 530 celebrities. The FGNet dataset is mainly used for testing age invariant face recognition, with 1002 face images from 82 persons.

We evaluated the proposed ATAM loss on FaceScrub of MegaFace Challenge 1, including both face identification and verification tasks. We followed the protocol of large training set as the training dataset contains more than 0.5M images, where the identities appearing in FaceScrub were removed from the training set. In addition, there are some noisy images from FaceScrub and MegaFace, hence we used the noises list proposed by~\cite{deng2019arcface} to clean it.

We employed an attribute predictor to predict the attributes for the MegaFace training set. In~\cite{11}, an ontology of 40 facial attributes are defined. We utilized the predicted attributes as an input in our proposed ATAM loss. 
For fair comparison, we implemented the Softmax, A-Softmax, CosFace, ArcFace, and our ATAM loss with the same architecture. Table~\ref{Table:face} shows the results of our models trained on the large protocol of MegaFace. The proposed model obtains the best performance on both identification and verification tasks, compared with related methods including SphereFace, ArcFace, and AdaptiveFace. This shows the effectiveness of the proposed ATAM loss through the final recognition rates on the MegaFace dataset.

\noindent\textbf{Experiments on YTF and LFW.}

We evaluated our proposed model on the
widely-used YTF~\cite{wolf2011face} and LFW~\cite{huang2008labeled} datasets. YTF contains 3,425 videos of 1,595 different persons downloaded from YouTube, with different variations of pose, illumination and expression, which is a popular dataset for unconstrained face recognition. In YTF, there are about 2.15 videos available for each person and a video clip has 181.3 frames on average. LFW~\cite{huang2008labeled} is a famous image dataset for face recognition, which contains 13,233 images from 5,749 different identities. The images are captured from the web in wild conditions, varying in pose, illumination, expression, age and background, leading to large intra-class variations.

\begin{table}[]
\centering
\small
\setlength\tabcolsep{2pt}
\begin{tabular}{c|cccc}
\hline
Method       & Training size & \#Models & LFW & YTF \\ \hline
Deep Face    &      4M         &    3     &  97.4   &   91.4  \\
FaceNet      &      200M         &   1      &  99.7   &  95.1   \\
DeepFR       &      2.6M         &  1       & 98.9     & 97.3   \\ 
DeepID2+     &     300K          &    25     &  99.5   &   93.2  \\
Center Face  &      0.7M         &    1     &  99.3   &    94.9  \\
Baidu        &   1.3M            &   1      &  99.1   &  -   \\
SphereFace   &     0.5M          &    1     &   99.4  &  95.0   \\
CosFace      &       5M$^*$        &     1    &   99.7  &  97.6   \\
UniformFace  &     6.1M          &     1    &  99.8   &   97.7  \\
AdaptiveFace &        5M       &     1    &   99.6  &  -   \\
\hline
Softmax      &     5M          &    1     &   98.8  & 95.7    \\
SphereFace   &      5M         &    1     &   99.6  & 96.6    \\
CosFace      &    5M           &    1     &   99.5  & 96.2    \\
ArcFace      &       5M        &     1    & 99.6    &  96.8   \\

ATAM         &    5M           &    1     &   99.7  &  97.9 
\end{tabular}
\caption{ Face verification (\%) on the LFW and YTF datasets.  “*” indicates although the dataset of CosFace contains 5M images, it is composed of several public datasets and a private face dataset, containing about more than 90K identities.}
\label{Table:LFW}

\end{table}

We followed the standard protocol of unrestricted with labeled outside data~\cite{huang2014labeled} to evaluate our model, and reported the result on the 5,000 and 6,000 pair testing images from YTF and LFW, respectively.  From the table~\ref{Table:LFW}, we observed that the usage of ATAM loss boosts the performance of 1.1\% on YTF compared to ArcFace with the same training data. It should be noted that the attributes which are passed to the proposed model during the training, are noisy due the error in the attribute prediction model. However, our proposed ATAM could exploit them to make the holistic feature space more discriminative and boost face recognition performance.

\subsection{Person re-identification:}\label{person-reidsection} 

Person re-identification (reID) goal is to spot the appearance of a same person in different scene. We evaluated our proposed method on two popular datasets, i.e., Market-1501~\cite{zheng2015scalable} and DukeMTMC-reID~\cite{zheng2017unlabeled}. Market-1501 contains 1,501 identities, 12,936 training images and 19,732 gallery images captured with 6 cameras. We utilized 27 attributes such as gender, hair length, carrying backpack, etc., annotated by~\cite{lin2019improving}. 
The DukeMTMC-reID dataset for re-ID has 1,812 identities from eight cameras. There are 1,404 identities appearing in more than two cameras and 408 identities (distractor ID) who appear in only one camera. We randomly picked 702 IDs as the training set and the remaining 702 IDs as the testing set. In the testing set, we select one query image for each ID in each camera and put the remaining images in the gallery. As a result, we get 16,522 training images with 702 identities, 2,228 query images of the other 702 identities, and 17,661 gallery images. We used 23 attributes such as gender, wearing hat, wearing boots, carrying backpack, etc., annotated by~\cite{lin2019improving}. 

We adopted two network architectures, i.e. a global
feature learning model backboned on ResNet-50 and a partfeature model named MGN~\cite{wang2018learning}. We used MGN due to its competitive performance and relatively concise structure. The original MGN uses a Softmax loss on each part feature branch for training. MGN~\cite{wang2018learning} is one of the state-of-the-art method which can learn multi-granularity part-level features. It utilizes both Softmax loss and triplet loss to facilitate a joint optimization. Following~\cite{sun2020circle}, we only used a single loss function in implementation of  “MGN (ResNet-50)+ ATAM loss” for simplicity. In addition, all the part features were concatenated into a single feature vector. We evaluate ATAM loss on re-ID task in Table~\ref{Table:re-id}.

\begin{table}[]
\centering
\small
\setlength\tabcolsep{2pt}
\begin{tabular}{ccccc}
\hline
\multirow{2}{*}{Method}                     & \multicolumn{2}{c}{Market-1501} & \multicolumn{2}{c}{DukeMTMC-reID} \\ \cline{2-5} 
                                            & R-1            & mAP            & R-1             & mAP             \\ \hline
\multicolumn{1}{c|}{PCB~\cite{sun2018beyond} (Softmax)}         & 93.8           & 81.6           & 83.3            & 69.2            \\
\multicolumn{1}{c|}{MGN~\cite{wang2018learning} (Softmax+Triplet)} & 95.7           & 86.9           & 88.7            & 78.4            \\
\multicolumn{1}{c|}{JDGL~\cite{zheng2019joint}}                   & 94.8           & 86.0           & 86.6            & 74.8            \\
\multicolumn{1}{c|}{APR~\cite{lin2019improving}}                    & 84.3           & 64.7           & 73.9            & 55.6            \\
\multicolumn{1}{c|}{AANet-50~\cite{tay2019aanet}}               & 93.9           & 82.5           & 86.4            & 72.6            \\
\multicolumn{1}{c|}{ResNet50 + AMSoftmax~\cite{wang2018additive}}   & 92.4           & 83.7           & 83.9            & 68.5            \\
\multicolumn{1}{c|}{ResNet50 + CircleLoss~\cite{sun2020circle}}  & 94.2           & 84.9           & -               & -               \\
\multicolumn{1}{c|}{ResNet50 + ATAM}        & 95.1           & 86.5           & 87.8            & 77.3            \\
\multicolumn{1}{c|}{MGN + AMSoftmax}        & 95.3           & 86.6           & 85.7            & 72.3            \\
\multicolumn{1}{c|}{MGN + CircleLoss}       & 96.1           & 87.4           & -               & -               \\
\multicolumn{1}{c|}{MGN + ATAM}             & 97.1           & 88.3           & 89.6            & 79.1           
\end{tabular}
\caption{ Evaluation of ATAM loss on re-ID task. We report
R-1 accuracy (\%) and mAP (\%).}
\label{Table:re-id}
\end{table}

We make following observations from Table~\ref{Table:re-id}. First, comparing ATAM loss against state-of-the-art, we find that ATAM loss achieves competitive re-ID accuracy, with a single loss function and without using auxiliary loss functions. Second, using privileged attributes, here also improves the accuracy of our proposed model which is consistent with the experimental results on face recognition task. Third, comparing ATAM loss with APR~\cite{lin2019improving} and AANet-50~\cite{tay2019aanet} methods which utilize attributes both in the training and inference time, we observe the superiority of ATAM loss.

\section{Conclusion}
Most existing methods aim to learn discriminative deep features, encouraging large inter-class distances and small intra-class variations. However,
they ignore the distribution of different classes in the holistic feature space, which may lead to severe locality and unbalance. Recent deep representation learning methods typically adopt class re-sampling or cost-sensitive learning schemes. In this paper, we proposed a novel approach which introduces the adaptive margins utilizing attributes to adaptively minimize intra-class variances. On two major deep feature learning tasks, i.e., face recognition and person re-identification,
our feature learning achieves performance on par with the
state-of-the-art. We believe that our approach could be very helpful for unbalanced data training in practice.

\bibliography{egbib}

\begin{thebibliography}{37}
\providecommand{\natexlab}[1]{#1}
\providecommand{\url}[1]{\texttt{#1}}
\expandafter\ifx\csname urlstyle\endcsname\relax
  \providecommand{\doi}[1]{doi: #1}\else
  \providecommand{\doi}{doi: \begingroup \urlstyle{rm}\Url}\fi

\bibitem[Ben-David et~al.(2010)Ben-David, Blitzer, Crammer, Kulesza, Pereira,
  and Vaughan]{4}
Shai Ben-David, John Blitzer, Koby Crammer, Alex Kulesza, Fernando Pereira, and
  Jennifer~Wortman Vaughan.
\newblock A theory of learning from different domains.
\newblock \emph{Machine learning}, 79\penalty0 (1-2):\penalty0 151--175, 2010.

\bibitem[Deng et~al.(2017)Deng, Zhou, and Zafeiriou]{deng2017marginal}
Jiankang Deng, Yuxiang Zhou, and Stefanos Zafeiriou.
\newblock Marginal loss for deep face recognition.
\newblock In \emph{Proceedings of the IEEE Conference on Computer Vision and
  Pattern Recognition Workshops}, pages 60--68, 2017.

\bibitem[Deng et~al.(2019)Deng, Guo, Xue, and Zafeiriou]{deng2019arcface}
Jiankang Deng, Jia Guo, Niannan Xue, and Stefanos Zafeiriou.
\newblock Arcface: Additive angular margin loss for deep face recognition.
\newblock In \emph{Proceedings of the IEEE Conference on Computer Vision and
  Pattern Recognition}, pages 4690--4699, 2019.

\bibitem[Duan et~al.(2019)Duan, Lu, and Zhou]{duan2019uniformface}
Yueqi Duan, Jiwen Lu, and Jie Zhou.
\newblock Uniformface: Learning deep equidistributed representation for face
  recognition.
\newblock In \emph{Proceedings of the IEEE Conference on Computer Vision and
  Pattern Recognition}, pages 3415--3424, 2019.

\bibitem[Feyereisl et~al.(2014)Feyereisl, Kwak, Son, and Han]{21}
Jan Feyereisl, Suha Kwak, Jeany Son, and Bohyung Han.
\newblock {Object localization based on structural SVM using privileged
  information}.
\newblock \emph{Advances in Neural Information Processing Systems (NIPS)},
  pages 208--216, 2014.

\bibitem[Fouad et~al.(2013)Fouad, Tino, Raychaudhury, and Schneider]{22}
Shereen Fouad, Peter Tino, Somak Raychaudhury, and Petra Schneider.
\newblock Incorporating privileged information through metric learning.
\newblock \emph{IEEE transactions on neural networks and learning systems},
  24\penalty0 (7):\penalty0 1086--1098, 2013.

\bibitem[Guo et~al.(2016)Guo, Zhang, Hu, He, and Gao]{guo2016ms}
Yandong Guo, Lei Zhang, Yuxiao Hu, Xiaodong He, and Jianfeng Gao.
\newblock Ms-celeb-1m: A dataset and benchmark for large-scale face
  recognition.
\newblock In \emph{European conference on computer vision}, pages 87--102.
  Springer, 2016.

\bibitem[He et~al.(2016)He, Zhang, Ren, and Sun]{31}
Kaiming He, Xiangyu Zhang, Shaoqing Ren, and Jian Sun.
\newblock Deep residual learning for image recognition.
\newblock \emph{IEEE Conference on Computer Vision and Pattern Recognition
  (CVPR)}, pages 770--778, 2016.

\bibitem[Hearst et~al.(1998)Hearst, Dumais, Osuna, Platt, and Scholkopf]{62}
Marti~A. Hearst, Susan~T Dumais, Edgar Osuna, John Platt, and Bernhard
  Scholkopf.
\newblock Support vector machines.
\newblock \emph{IEEE Intelligent Systems and their applications}, 13\penalty0
  (4):\penalty0 18--28, 1998.

\bibitem[Huang and Learned-Miller(2014)]{huang2014labeled}
Gary~B Huang and Erik Learned-Miller.
\newblock Labeled faces in the wild: Updates and new reporting procedures.
\newblock \emph{Dept. Comput. Sci., Univ. Massachusetts Amherst, Amherst, MA,
  USA, Tech. Rep}, pages 14--003, 2014.

\bibitem[Huang et~al.(2008)Huang, Mattar, Berg, and
  Learned-Miller]{huang2008labeled}
Gary~B Huang, Marwan Mattar, Tamara Berg, and Eric Learned-Miller.
\newblock Labeled faces in the wild: A database forstudying face recognition in
  unconstrained environments.
\newblock 2008.

\bibitem[Kemelmacher-Shlizerman et~al.(2016)Kemelmacher-Shlizerman, Seitz,
  Miller, and Brossard]{kemelmacher2016megaface}
Ira Kemelmacher-Shlizerman, Steven~M Seitz, Daniel Miller, and Evan Brossard.
\newblock The megaface benchmark: 1 million faces for recognition at scale.
\newblock In \emph{Proceedings of the IEEE conference on computer vision and
  pattern recognition}, pages 4873--4882, 2016.

\bibitem[Lin et~al.(2019)Lin, Zheng, Zheng, Wu, Hu, Yan, and
  Yang]{lin2019improving}
Yutian Lin, Liang Zheng, Zhedong Zheng, Yu~Wu, Zhilan Hu, Chenggang Yan, and
  Yi~Yang.
\newblock Improving person re-identification by attribute and identity
  learning.
\newblock \emph{Pattern Recognition}, 95:\penalty0 151--161, 2019.

\bibitem[Liu et~al.(2019)Liu, Zhu, Lei, and Li]{liu2019adaptiveface}
Hao Liu, Xiangyu Zhu, Zhen Lei, and Stan~Z Li.
\newblock Adaptiveface: Adaptive margin and sampling for face recognition.
\newblock In \emph{Proceedings of the IEEE Conference on Computer Vision and
  Pattern Recognition}, pages 11947--11956, 2019.

\bibitem[Liu et~al.(2016)Liu, Wen, Yu, and Yang]{liu2016large}
Weiyang Liu, Yandong Wen, Zhiding Yu, and Meng Yang.
\newblock Large-margin softmax loss for convolutional neural networks.
\newblock In \emph{ICML}, volume~2, page~7, 2016.

\bibitem[Liu et~al.(2017)Liu, Wen, Yu, Li, Raj, and Song]{liu2017sphereface}
Weiyang Liu, Yandong Wen, Zhiding Yu, Ming Li, Bhiksha Raj, and Le~Song.
\newblock Sphereface: Deep hypersphere embedding for face recognition.
\newblock In \emph{Proceedings of the IEEE conference on computer vision and
  pattern recognition}, pages 212--220, 2017.

\bibitem[Liu et~al.(2015)Liu, Luo, Wang, and Tang]{11}
Ziwei Liu, Ping Luo, Xiaogang Wang, and Xiaoou Tang.
\newblock Deep learning face attributes in the wild.
\newblock \emph{IEEE International Conference on Computer Vision (ICCV)}, pages
  3730--3738, 2015.

\bibitem[Motiian et~al.(2016)Motiian, Piccirilli, Adjeroh, and Doretto]{3}
Saeid Motiian, Marco Piccirilli, Donald~A Adjeroh, and Gianfranco Doretto.
\newblock Information bottleneck learning using privileged information for
  visual recognition.
\newblock In \emph{IEEE Conference on Computer Vision and Pattern Recognition
  (CVPR)}, 2016.

\bibitem[Ng and Winkler(2014)]{ng2014data}
Hong-Wei Ng and Stefan Winkler.
\newblock A data-driven approach to cleaning large face datasets.
\newblock In \emph{2014 IEEE international conference on image processing
  (ICIP)}, pages 343--347. IEEE, 2014.

\bibitem[Saenko et~al.(2010)Saenko, Kulis, Fritz, and Darrell]{5}
Kate Saenko, Brian Kulis, Mario Fritz, and Trevor Darrell.
\newblock Adapting visual category models to new domains.
\newblock \emph{European conference on computer vision}, pages 213--226, 2010.

\bibitem[Schroff et~al.(2015)Schroff, Kalenichenko, and
  Philbin]{schroff2015facenet}
Florian Schroff, Dmitry Kalenichenko, and James Philbin.
\newblock Facenet: A unified embedding for face recognition and clustering.
\newblock In \emph{Proceedings of the IEEE conference on computer vision and
  pattern recognition}, pages 815--823, 2015.

\bibitem[Sharmanska et~al.(2013)Sharmanska, Quadrianto, and Lampert]{10}
Viktoriia Sharmanska, Novi Quadrianto, and Christoph~H Lampert.
\newblock Learning to rank using privileged information.
\newblock \emph{IEEE International Conference on Computer Vision (ICCV)}, pages
  825--832, 2013.

\bibitem[Sun et~al.(2014{\natexlab{a}})Sun, Chen, Wang, and Tang]{sun2014deep}
Yi~Sun, Yuheng Chen, Xiaogang Wang, and Xiaoou Tang.
\newblock Deep learning face representation by joint
  identification-verification.
\newblock In \emph{Advances in neural information processing systems}, pages
  1988--1996, 2014{\natexlab{a}}.

\bibitem[Sun et~al.(2014{\natexlab{b}})Sun, Wang, and Tang]{34}
Yi~Sun, Xiaogang Wang, and Xiaoou Tang.
\newblock Deep learning face representation from predicting 10,000 classes.
\newblock In \emph{Proceedings of the IEEE Conference on Computer Vision and
  Pattern Recognition (CVPR)}, pages 1891--1898, 2014{\natexlab{b}}.

\bibitem[Sun et~al.(2018)Sun, Zheng, Yang, Tian, and Wang]{sun2018beyond}
Yifan Sun, Liang Zheng, Yi~Yang, Qi~Tian, and Shengjin Wang.
\newblock Beyond part models: Person retrieval with refined part pooling (and a
  strong convolutional baseline).
\newblock In \emph{Proceedings of the European Conference on Computer Vision
  (ECCV)}, pages 480--496, 2018.

\bibitem[Sun et~al.(2020)Sun, Cheng, Zhang, Zhang, Zheng, Wang, and
  Wei]{sun2020circle}
Yifan Sun, Changmao Cheng, Yuhan Zhang, Chi Zhang, Liang Zheng, Zhongdao Wang,
  and Yichen Wei.
\newblock Circle loss: A unified perspective of pair similarity optimization.
\newblock \emph{arXiv preprint arXiv:2002.10857}, 2020.

\bibitem[Tay et~al.(2019)Tay, Roy, and Yap]{tay2019aanet}
Chiat-Pin Tay, Sharmili Roy, and Kim-Hui Yap.
\newblock Aanet: Attribute attention network for person re-identifications.
\newblock In \emph{Proceedings of the IEEE Conference on Computer Vision and
  Pattern Recognition}, pages 7134--7143, 2019.

\bibitem[Vapnik and Vashist(2009)]{9}
Vladimir Vapnik and Akshay Vashist.
\newblock A new learning paradigm: Learning using privileged information.
\newblock \emph{Neural networks}, 22\penalty0 (5):\penalty0 544--557, 2009.

\bibitem[Wang et~al.(2018{\natexlab{a}})Wang, Cheng, Liu, and
  Liu]{wang2018additive}
Feng Wang, Jian Cheng, Weiyang Liu, and Haijun Liu.
\newblock Additive margin softmax for face verification.
\newblock \emph{IEEE Signal Processing Letters}, 25\penalty0 (7):\penalty0
  926--930, 2018{\natexlab{a}}.

\bibitem[Wang et~al.(2018{\natexlab{b}})Wang, Yuan, Chen, Li, and
  Zhou]{wang2018learning}
Guanshuo Wang, Yufeng Yuan, Xiong Chen, Jiwei Li, and Xi~Zhou.
\newblock Learning discriminative features with multiple granularities for
  person re-identification.
\newblock In \emph{Proceedings of the 26th ACM international conference on
  Multimedia}, pages 274--282, 2018{\natexlab{b}}.

\bibitem[Wang et~al.(2018{\natexlab{c}})Wang, Wang, Zhou, Ji, Gong, Zhou, Li,
  and Liu]{wang2018cosface}
Hao Wang, Yitong Wang, Zheng Zhou, Xing Ji, Dihong Gong, Jingchao Zhou, Zhifeng
  Li, and Wei Liu.
\newblock Cosface: Large margin cosine loss for deep face recognition.
\newblock In \emph{Proceedings of the IEEE Conference on Computer Vision and
  Pattern Recognition}, pages 5265--5274, 2018{\natexlab{c}}.

\bibitem[Wen et~al.(2016)Wen, Zhang, Li, and Qiao]{wen2016discriminative}
Yandong Wen, Kaipeng Zhang, Zhifeng Li, and Yu~Qiao.
\newblock A discriminative feature learning approach for deep face recognition.
\newblock In \emph{European conference on computer vision}, pages 499--515.
  Springer, 2016.

\bibitem[Wolf et~al.(2011)Wolf, Hassner, and Maoz]{wolf2011face}
Lior Wolf, Tal Hassner, and Itay Maoz.
\newblock Face recognition in unconstrained videos with matched background
  similarity.
\newblock In \emph{CVPR 2011}, pages 529--534. IEEE, 2011.

\bibitem[Zhang et~al.(2016)Zhang, Zhang, Li, and Qiao]{zhang2016joint}
Kaipeng Zhang, Zhanpeng Zhang, Zhifeng Li, and Yu~Qiao.
\newblock Joint face detection and alignment using multitask cascaded
  convolutional networks.
\newblock \emph{IEEE Signal Processing Letters}, 23\penalty0 (10):\penalty0
  1499--1503, 2016.

\bibitem[Zheng et~al.(2015)Zheng, Shen, Tian, Wang, Wang, and
  Tian]{zheng2015scalable}
Liang Zheng, Liyue Shen, Lu~Tian, Shengjin Wang, Jingdong Wang, and Qi~Tian.
\newblock Scalable person re-identification: A benchmark.
\newblock In \emph{Proceedings of the IEEE international conference on computer
  vision}, pages 1116--1124, 2015.

\bibitem[Zheng et~al.(2017)Zheng, Zheng, and Yang]{zheng2017unlabeled}
Zhedong Zheng, Liang Zheng, and Yi~Yang.
\newblock Unlabeled samples generated by gan improve the person
  re-identification baseline in vitro.
\newblock In \emph{Proceedings of the IEEE International Conference on Computer
  Vision}, pages 3754--3762, 2017.

\bibitem[Zheng et~al.(2019)Zheng, Yang, Yu, Zheng, Yang, and
  Kautz]{zheng2019joint}
Zhedong Zheng, Xiaodong Yang, Zhiding Yu, Liang Zheng, Yi~Yang, and Jan Kautz.
\newblock Joint discriminative and generative learning for person
  re-identification.
\newblock In \emph{Proceedings of the IEEE Conference on Computer Vision and
  Pattern Recognition}, pages 2138--2147, 2019.

\end{thebibliography}

\typeout{get arXiv to do 4 passes: Label(s) may have changed. Rerun} 

\end{document}